\documentclass[letterpaper, 10 pt, conference]{ieeeconf}  % Comment this line out if you need a4paper

\IEEEoverridecommandlockouts                              % This command is only needed if 
\usepackage{graphics} % for pdf, bitmapped graphics files
\usepackage{epsfig} % for postscript graphics files
\usepackage{mathptmx} % assumes new font selection scheme installed
\usepackage{times} % assumes new font selection scheme installed
\usepackage{amsmath} % assumes amsmath package installed
\usepackage{amssymb}  % assumes amsmath package installed

\usepackage{flushend}

\usepackage{cite}

\title{\LARGE \bf
Learning Robust Grasping Strategy Through Tactile Sensing and Adaption Skill
}

\author{Yueming Hu$^{1}$, Mengde Li$^{2,4}$, Songhua Yang$^{3}$, Xuetao Li$^{3}$, Sheng Liu$^{1}$,~\IEEEmembership{Fellow, IEEE} \\
and Miao Li$^{2,4,*}$,~\IEEEmembership{Senior Member, IEEE}
\thanks{$^{1}$ School of Power and Mechanical Engineering, Wuhan University, Hubei,China}
\thanks{$^{2}$ School of Technological Sciences, Wuhan University, Hubei, China }
\thanks{$^{3}$ School of Computer Science, Wuhan University, Hubei,China}
\thanks{$^{4}$ School of Microelectronics, Wuhan University, Hubei, China}
\thanks{$^{*}$ Corresponding author.(e-mail: miao.li@whu.edu.cn)}
}

\begin{document}

\maketitle
\pagestyle{empty}
\raggedend

%%%%%%%%%%%%%%%%%%%%%%%%%%%%%%%%%%%%%%%%%%%%%%%%%%%%%%%%%%%%%%%%%%%%%%%%%%%%%%%%
\begin{abstract}

Robust grasping represents an essential task in robotics, necessitating tactile feedback and reactive grasping adjustments for robust grasping of objects. Previous research has extensively combined tactile sensing with grasping, primarily relying on rule-based approaches, frequently neglecting post-grasping difficulties such as external disruptions or inherent uncertainties of the object's physics and geometry. To address these limitations, this paper introduces an human-demonstration-based adaptive grasping policy base on tactile, which aims to achieve robust gripping while resisting disturbances to maintain grasp stability. Our trained model generalizes to daily objects with seven different sizes, shapes, and textures. Experimental results demonstrate that our method performs well in dynamic and force interaction tasks and exhibits excellent generalization ability.

\end{abstract}

%%%%%%%%%%%%%%%%%%%%%%%%%%%%%%%%%%%%%%%%%%%%%%%%%%%%%%%%%%%%%%%%%%%%%%%%%%%%%%%%
\section{INTRODUCTION}

The grasping operation is a fundamental action commonly performed by robots \cite{thrun2002probabilistic}. Due to its versatility, it finds application in various fields such as warehousing \cite{ren2023visual}, food processing \cite{kiyokawa2019generation}, rescue operations \cite{hatano2017estimation}, aerospace \cite{xu2021optimal}, and agriculture \cite{pettersson2010design}. During the grasping process, factors such as the shape, state, material of the object being grasped, as well as the external environment, need to be considered. For example, when a robot holds a power drill to perform assembly tasks in a workshop, the high-frequency vibrations of the drill could cause it to slip out of the gripper. Therefore, exploring adaptive adjustment strategies for robotic grasping operations is crucial, ensuring robustness and stability even when faced with numerous external disturbances.

Robots generally lack prior knowledge about objects, making the stability of the grasping process susceptible to issues such as sliding vibrations or changes in mass \cite{li2023ptfd}. For example, when a robot uses one arm to grasp a cup while the other fills it with a beverage, changes in the object's mass or weight exceeding the estimated value can alter the grasping configuration, thus requiring a stable immediate response. 

\begin{figure}
\centering
\includegraphics[width=\columnwidth]{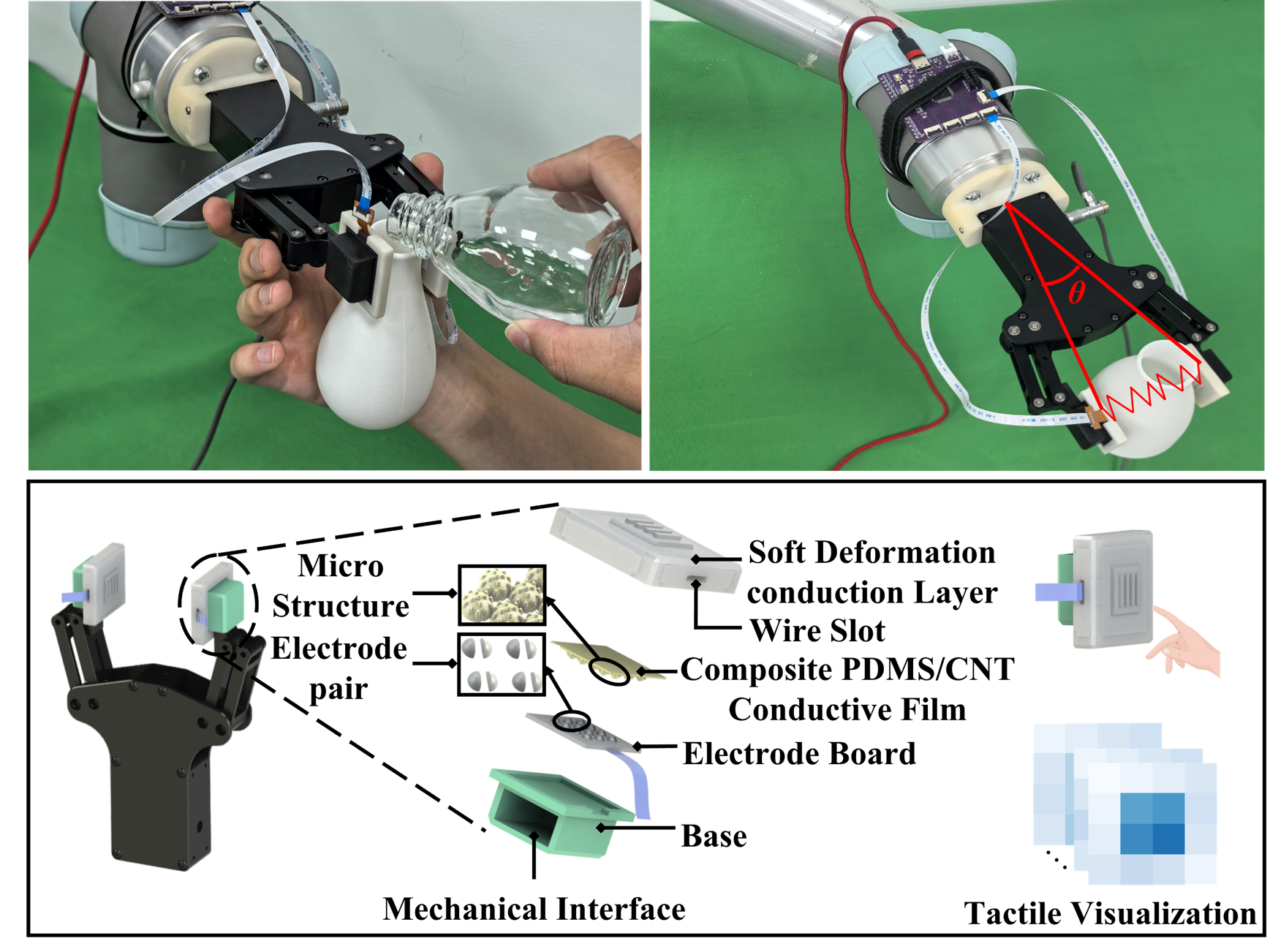}
\caption{Human demonstration and Flexible tactile sensor. A human controls the gripper manually to change the parameter $\theta$ as shown in the figure, with with adding water as an external disturbance. This figure details the mechanical structure and tactile visualization of the tactile sensor.} 
\label{fig：Fig.1}
\end{figure}

To perceive and understand these external disturbances during the grasping process, tactile sensors are indispensable. Previous research has extensively explored the role of tactile perception in robust grasping and has proposed various methods to enhance the robustness and stability \cite{qiao2019self}, \cite{xu2021adagrasp}, \cite{dexin2023high}. Specifically, these studies utilizing tactile feedback mainly focus on uncertainties originating from the object's geometric model \cite{chen2023sliding}, position \cite{chen2015adaptive}, and external disturbances \cite{khadivar2023adaptive}. These uncertainties can directly affect the relative configuration between the robotic hand and the object, thus directly impacting the grasping stability.

For example, an object-level impedance control strategy was introduced in \cite{lilearning}, proposing a robust method to handle grasping uncertainties. Even when a person applies arbitrary pulling actions on the object being grasped, the Allegro hand can continuously adjust its state through tactile feedback to maintain a stable grasp. The initial grasp in this method is pre-defined rather than learned, which may reduce its robustness in complex situations. In \cite{hyttinen2017estimating}, a method of simulating tactile data near the grasp position to evaluate and correct the grasp of new objects is proposed. However, since only a single corrective action is performed, the robotic hand may accidentally knock over the object, preventing re-establishment of contact and thus failing to find additional adaptive measures, which leaves potential for grasping failure. In \cite{kolamuri2021improving}, a method for detecting rotational displacement of objects caused by non-center-of-mass grasping is proposed. Researchers, utilizing the visual-tactile information provided by GelSight, integrate the rotational detection feedback into a closed-loop re-grasping framework to robots. A haptic event-based grasping algorithm is proposed in \cite{kim2020tactile}, which uses predefined events as state transition triggers, inspired by human grasping behavior (hold, place, unload). This means that each step of the system is predefined, and transitions between states are triggered by specific haptic events, which limits its adaptability in complex and dynamic environments. To sum up, these previous studies have largely focused on predefined and logic-based controllers and have concentrated more on single adjustment actions rather than continuous actions.

To this end, in this paper we present an human-demonstration-based adaptive grasping policy for resisting any external disturbances (human interface, vibration, deformation) and validated on an electric gripper. Specifically, first, we collect stable grasp data by having humans manually control the gripper. Next, for the adaptive strategy, we design an initial grasp generator, a stability estimator, and an adaptive grasp strategy. To begin with, an initial grasp action is generated by the grasp generator. Essentially, the stability estimator functions as a probabilistic classifier, where an output below a given threshold indicates an unstable grasp. Once the grasp adapter proves the current grasp to be unstable, it adjusts using parameters learned from skilled demonstrations. Finally, we train the adaptive strategy using a neural network based on self-attention mechanisms. Further, we conduct grasp stability tests on seven different objects. Experimental results show that the strategy successfully handles arbitrary dragging by humans and disturbances such as adding water to containers, demonstrating the method's potential in robotic grasping. Briefly, our contributions can be summarized as follows: 

1) A grasp generator and a stability estimator for the adaptive grasping process is presented.

2) A real-time dynamic adjustment strategy for robots based on human demonstrations is proposed.

The reminder of this paper is organized as follows: The modeling and analysis of the proposed adaptive grasping policy are shown in Section II. The experimental results are presented in Section III to demonstrate the effectiveness of this method, followed by a conclusion in Section IV, with a discussion on the limits of the current design.

\section{System Design and Implementation}
In this section, we will introduce our method for learning stability estimator and grasp adaption, relying exclusively on environmental insights provided by tactile sensors integrated into the robotic gripper. This method signifies a departure from conventional methodologies that heavily depend on vision, instead focusing on tactile sensitivity to navigate the complexities of object manipulation.

Our method can be roughly segmented to three parts: a initial grasp generator, a stability estimator and a grasp adaption strategy. The initial grasp devised by the generator apply a minimal grasping force to ensure object stability. Once the object is lifted, the stability estimator assesses the current stability by analyzing tactile reading and joint angle. If the present grasp is proved unstable, the adaption strategy generates and executes corrective actions. Our approach involves applying these two methods iteratively until the grasp remains stable or there is no more external interference. Our research concentrates on strategies for adaptive adaptation following the initial grasp, highlighting the capacity to resist external disruptions or internal interference.

\subsection{Initial Grasp Generator}
The concept of initial grasp generator is a generator learned from experience. First, We collect grasp examples the object to learn the parallel gripper's configuration, including grasping force, angle, and wrist pose. Once the gripper are artificially preformed, it will proceed local exploration in the vicinity of contact points, triggered by the tactile sensors' response. For each example, along with its corresponding control parameters, will form a grasp configuration dataset and be used to suggest initial grasps for novel objects. Unlike the generator presented by Detry et al. \cite{detry2013learning}, our approach utilizes tactile feedback instead of visual sense.

The first step involves detection of object surface to identify potential contact areas. In essence, our aim is to identify the geometric properties on the objects' surface, encompassing the area and size of interaction zones. Secondly, for each potential grasp area, investigating the contact patterns of the area is essential. The initial grasp is then executed using the grasp parameters encoded in the chosen initial grasp template, which composes the dataset for initial grasp, instructing the parallel gripper on finger placement, the required force, and orientation for a stable grip. The Initial Grasp Templates are the cornerstone of our tactile-based grasp planning system, enabling the robot to make informed predictions on its interactions with unseen objects. To be precise, we employed a supervised learning method namely behavior cloning to bypass the direct evaluation of initial grasp configuration, and instead map the template to grasp vector $a\in \mathbb{R}$ for the gripper. In this way, we are able to output the optimal initial grasp.

Let $D_{gp} = \left\{ \left(S_i, \theta_i\right) \right\}_{i=1}^{N}$ denote the dataset with $N$ observations for the initial grasp. Among them, $S_i = \left(S_1^i, S_2^i\right) \in \mathbb{R}^{32}$ denotes the tactile reading, while $S_1^i, S_2^i \in \mathbb{R}^{16}$ is the tactile reading from each finger. $a_i \in \mathbb{R}$ indicates the angle from the hand. Our model for initial grasp can be expressed as:

\begin{equation}
w *=\underset{w}{\arg \min } E_{(s, a) B}\left[L\left(\pi_w(s), a\right)\right]
\end{equation}

where $w$ is the weight of our neural network model. $(s, a)$ is a state-action pair, with $s=(S, \theta)$ denoting current state and $a$ indicating the gripper's angle taken in state $s$ and $\hat{a}_{i}=\pi\left(s;w\right)$ represents the policy function based on the parameter $w$.

\begin{equation}
    w_{i+1}\leftarrow w_i-\alpha\nabla_w\sum_iL(\pi_\theta(s_i),a_i|(s_i,a_i)\in D_{gp})
\label{back}
\end{equation}

where $\alpha$ and $L$ denote the learning rate and the loss function. Detaily, we choose the mean squared error as the loss function in (\ref{loss}). Ultimately, through the backpropagation in (\ref{back}), we are able to adjust the weights $w$ to minimize the loss function values on the training set, thereby obtaining the optimal neural network model. This model will be able to predict initial grasp $a$ from the given tactile readings and grasp angles. 

\begin{equation}
    L(s,a;w)=\frac{1}{2N}\sum_{i=1}^{N}(\hat{a}_{i}-a_{i})^2
\label{loss}
\end{equation}

where $a$ and $\pi(s;w)$ denote the initial grasp actions (angle) from the training set and the initial grasp actions output by the model, respectively.

\subsection{Stability Estimator}

Given an initial grasp $a$ from $\text{II-A}$, real-time analysis of stability is essential to maintain the grasp. Our stability estimator integrates tactile readings with joint angle to predict whether the current grasp is stable or unstable. Tactile cues might be a good indication of stability, e.g. a tactile pad with no registered contacts typically signals an unstable grasp. Different from previous approach  In this paper, we define grasp stability estimation as a binary classification problem and train our model with Gaussian Mixture Model (GMM). We define grasp template mentioned in $\text{II-A}$ as $\Omega$. For stability estimation, we define the current grasp as $X_{*}$ = ($S_{*}$, $\theta_{*}$, $P_{*}$), where $P\in \mathbb{R}^{7}$ represents the location of the end effector of the robotic arm. The likelihood of a grasp $X_{*}$ within a GMM framework denoted by $\Omega$ with $m$ Gaussian components is as follows:

\begin{equation}
    p(X_*|\Omega)=\sum_{i=1}^m\pi_i\mathcal{N}(X_*|\mu_i,\Sigma_i)
\label{equation3}
\end{equation}

In (\ref{equation3}), $\pi_i$ denotes the probability distribution of the latent variables. ${N}(X_*|\mu_i,\Sigma_i)$ is the Gaussian distribution that features a mean $\mu_i$ and a covariance $\Sigma_i$ as follows.

\begin{equation}
\mu_i=\begin{bmatrix}\mu_{S,i}\\\mu_{\theta,i}\\\mu_{P,i}\end{bmatrix},\Sigma_i=\begin{bmatrix}\Sigma_{SS,i}&\Sigma_{S\theta,i}&\Sigma_{SP,i}\\\Sigma_{{\theta}S,i}&\Sigma_{{\theta}{\theta},i}&\Sigma_{{\theta}P,i}\\\Sigma_{PS,i}&\Sigma_{P\theta,i}&\Sigma_{PP,i}\end{bmatrix}
\end{equation}

 When the likelihood exceeds a given threshold $T_{h}$ e.i. $p(X_{*}) > te$, the grasp is proved stable. The selection of the discrimination threshold $te$ within the range $[a,b]$, is based on the ROC   (Receiver Operating Characteristic) curve, with $a$ and $b$ corresponding the minimal and maximal likelihood of each Gaussian component at two standard deviations, respectively, are:

\begin{align}
Lik_{2\sigma}(i) &= (2\pi)^{-\frac{d}{2}}|\Sigma_{i}|^{-\frac{1}{2}}e^{-2} \nonumber\\
a &= \min_{i=1\ldots m} Lik_{2\sigma}(i) && \label{eq:example}\\
b &= \max_{i=1\ldots m} Lik_{2\sigma}(i) \nonumber
\end{align}

\begin{figure}
\centering
\includegraphics[width=\columnwidth]{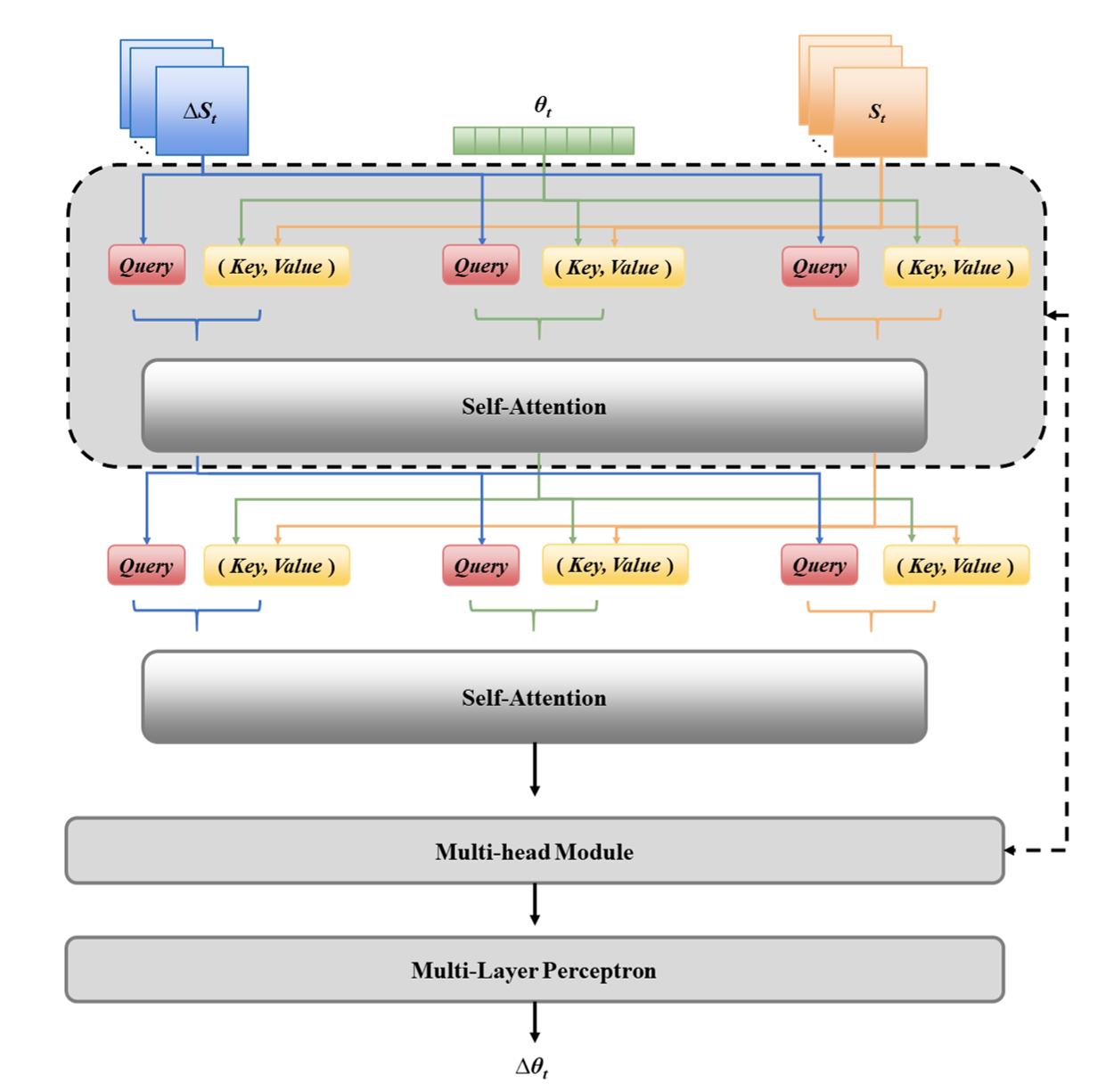}
\caption{Schematic diagram about the self-attention mechanism for learning the adaptive grasping skills.} 
\label{network}
\end{figure}

\begin{figure*}
\centering 
\includegraphics[width=\textwidth]{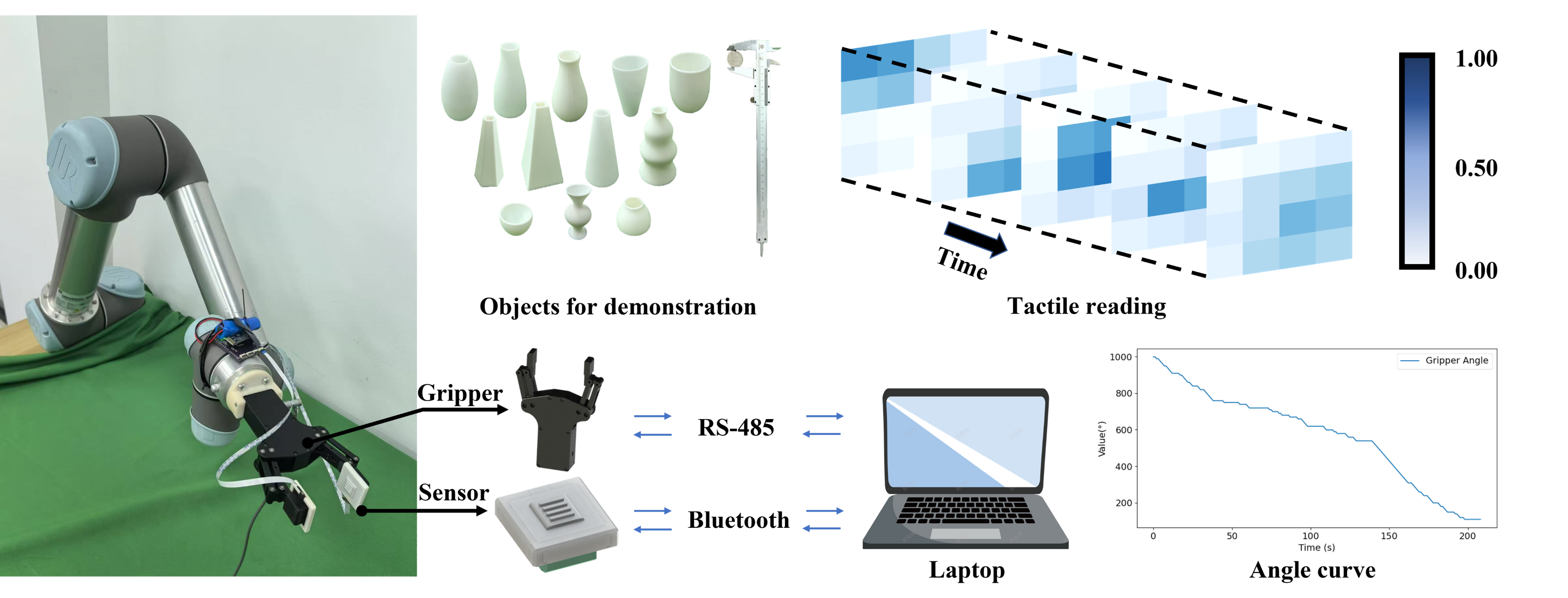} 
\caption{The experimental setup for adaptive grasping demonstrations. Both fingers of the electric parallel gripper are equipped with tactile sensors. We utilized a printed circuit board to receive tactile signals through a Bluetooth, forwarding them to a laptop. The gripper interacts with the laptop through the RS485 port and transmits its angle to the host computer through the USB port.} 
\label{experimentsetup} 
\end{figure*} 

\subsection{Grasp Adaption}

Once the stability proposed in $\text{III-B}$ inform us the current grasp is unstable, a grasp adaption method is essential. This component is designed to dynamically adjust the grasp configuration based on tactile feedback and object properties. To be specific, our method is to adjust the current grasp configuration within the scope of the initial grasp, that is, applying a local exploration policy to find potential corrective actions $\alpha$. Due to the underactuated hand, namely, the parallel gripper, we choose to only consider the actions of the gripper, e.i. $\alpha$ can be described as:

\begin{equation}
    \alpha=\Delta\theta 
\label{equation6}
\end{equation}
where $\Delta\theta$ is the change in the angle of the gripper. Given the current angle $\theta$, the updated angle $\theta^{\prime}$ after adjustment can be calculated as:

\begin{equation}
    \theta'=\theta+\alpha 
\end{equation}

We define our model as an human-demonstration-based adaptive grasp policy based on tactile(he-AGPT). As shown in (\ref{equation6}), our policy is to adapt the angle $\theta$. This method is used to indirectly regulate the forces and contact areas of both fingers. Once an uncertainty is detected such as vibrations on fingertips, the desired grasp angle $\hat{\theta}$ is predicted from model, given changes in tactile reading and the current angle $\theta$. Let us denote the dataset by:

\begin{equation}
    D_{(ga)}=\left\{\left(\Delta S_{i}, S_{i}, \theta_{i},\Delta\theta_{i}\right)\right\}_{i=1...N}
\end{equation}
Adaptation is a dynamic process of adjustment, so we apply both ${\Delta}S$ and $S$ to monitor the state of the grasp. 

Since tactile readings have time sequential characteristics that can be handled effectively by the self-attention mechanism \cite{vaswani2017attention}, and output adaptive actions as a sequence of time-continuous movements makes the adjustments smoother, we consider using self-attention layer for our model. Mathematically, the self-attention mechanism is expressed as follows:
\begin{equation}
    \text{Attention}(Q,K,V)=\text{softmax}(\frac{QK^T}{\sqrt{d_k}})V
\label{equation9}
\end{equation}

In (\ref{equation9}), $Q$, $K$ and $V$ are all obtained by different linear transformations of the input data $X_{i}$ = ($\Delta S_{i}$, $\Delta \theta_{i}$), and $W_{Q}$, $W_{K}$ and $W_{V}$ are the corresponding weight matrices. As illustrated in Fig. \ref{network}, self-attention layers are applied iteratively and then followed by a Multilayer Perceptron (MLP) to produce the final output $\Delta \theta_{t}$.

\section{Experiments}
In this section, we present an experiment on a real robot arm platform, demonstrating the feasibility of our adaptive grasping approach. We first introduce the setup of our experiment platform in $\text{III-A}$. Subsequently, expert demonstrations are utilised to data collection, which will be detailed in $\text{III-B}$. The results of the grasp adaption on novel objects are presented in $\text{III-C}$. The sensor used in this experiment, as shown in Fig. \ref{fig：Fig.1}, is a novel piezoresistive sensor designed based on previous work \cite{zhao2023novel}, \cite{lei2022biomimetic}. \cite{zhao2023novel} presented a soft tactile palm, composed of an electrode array, conductive foam, and elastic skin, detects contact locations and forces. Experiments demonstrate clear tactile responses for various everyday objects and an estimation error of 0.38 $N$ in normal contact force. In \cite{lei2022biomimetic}, the paper introduces a bionic tactile palm using a rigid circuit board embedded in elastic silicone rubber with an electrode array and conductive liquid. It measures impedance changes to detect contact locations and forces. Experiments show it estimates contact points with 2.3 millimeter average precision, a 4.3 millimeter maximum error, and a 0.38 Newton root mean square error in normal force.

\subsection{Experiment Setup} 

The demonstration platform is shown in Fig. \ref{experimentsetup}. In this experiment, a collaborative robot arm was used as an actuator, equipped with a signal acquisition module to establish communication between tactile sensors and the laptop. We apply a 2 DOFS (degree of freedom) \textit{EG2-4B} electric gripper from Inspire-robots\footnote{https://www.inspire-robots.com} with two fingers, and both fingertips are equipped with tactile sensors as mentioned in Fig. \ref{fig：Fig.1}. Each sensor consists of an array of 16 spatial electrodes, so the tactile data has a dimension of 32, which features contact force, contact position and deformation. The gripper and tactile sensor communicate with the host computer via RS-485 bus and serial port, respectively. The sensory information was recorded and processed by a laptop with Intel i7 CPU and Nvidia GTX 1650ti GPU, with an environment of Ubuntu20.04.6 LTS and PyTorch 2.3.0.

\begin{figure}
\centering
\includegraphics[width=\columnwidth]{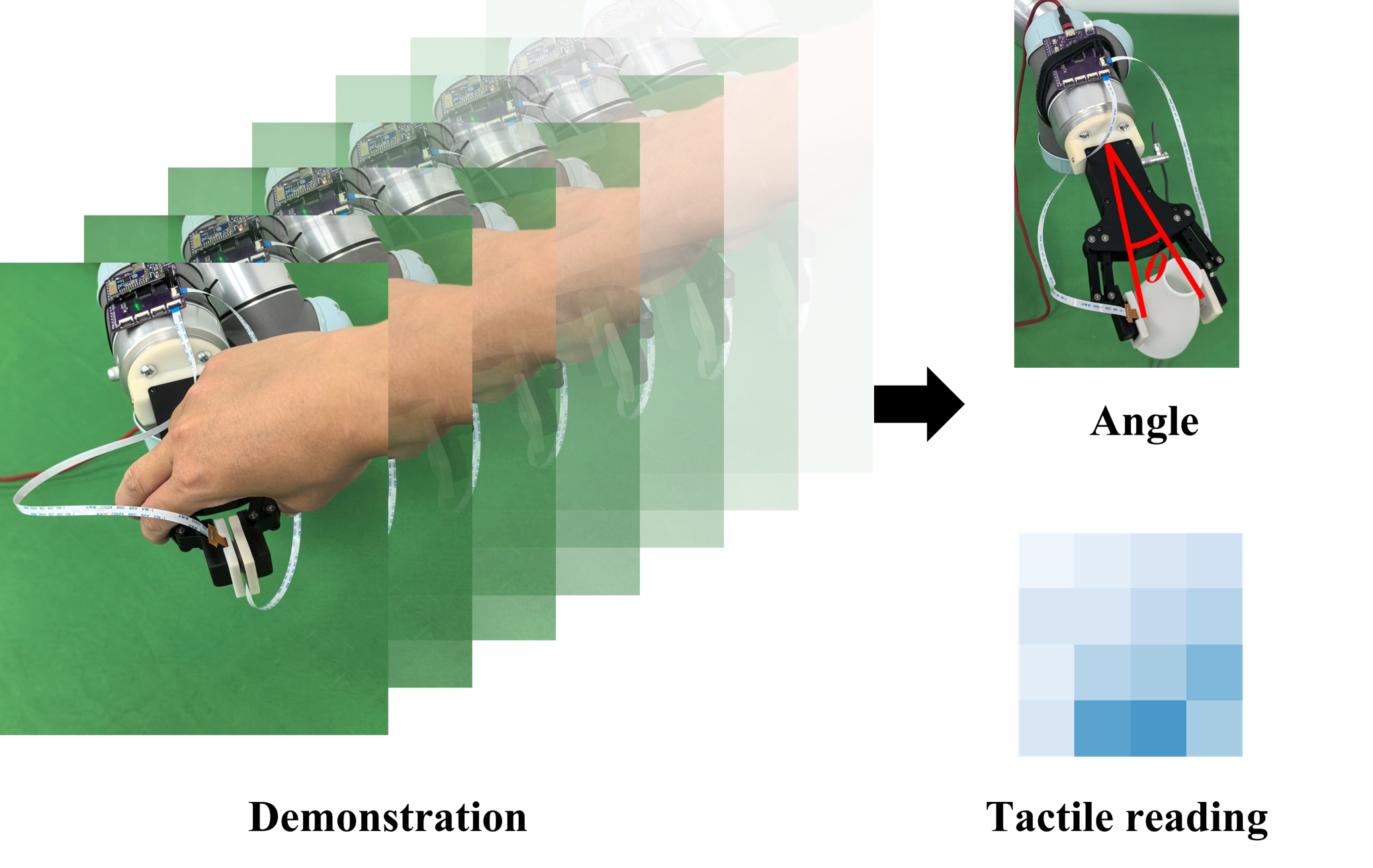}
\caption{Expert demonstrations. Experts demonstrate adaptive grasping with different forces and controlled gripper angles, thus recording the corresponding tactile readings and angles.} 
\label{expertdemonstration}
\end{figure}

\subsection{Data Collection}

During data acquisition, the gripper is in demonstration mode, allowing the operator to manually control its opening and closing continuously as shown in Fig. \ref{expertdemonstration}. We set our data sampling frequency to 160Hz. It is sufficient to implement the grasp once and continuously apply disturbances, rather than repeatedly grasping objects. 

For the initial grasp generator, experts demonstrate by manually controlling the gripper, specifically recording data from the moment the gripper first contacts the object until it can stably hold the object with minimal force. For the data used in stability assessment, similar to the positive samples in the generator, negative samples consist of data from the moment the gripper first contacts the object until the object drops. In grasp adaption, We employed two types of external disturbances: one is manual interference, where a person applies pull or pressure to the object, and the other is by adding water to increase the object’s weight. During demonstration, additional clamping is needed to enhance the contact force between the object and sensor. Information is gathered from various points on the objects during the grasping process. Tactile sensors collect data $S$ and record the gripper angle $\theta$. In total, 1200 experiments were performed on 12 objects, considering physical attributes, surface curvature features, and initial gripping positions. Gripper's angles depend on the object's geometry, while stiffness dictates the tactile reading range. Ultimately, a total of 30000 (10000+6000+14000) data points were collected for the training set.

\subsection{Model Training}

\begin{figure}
\centering
\includegraphics[width=\columnwidth]{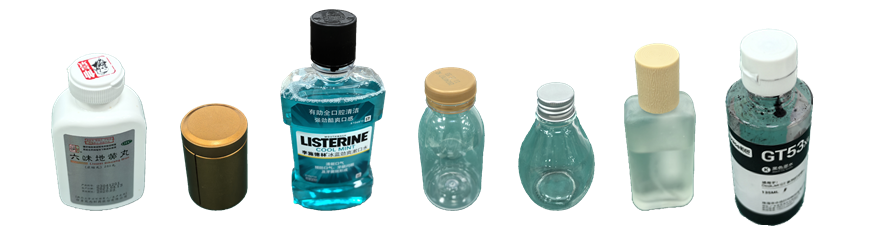}
\caption{Objects for testing: pill box, tea can, mouthwash, milk bottle, wine bottle, perfume, ink.} 
\label{testitems}
\end{figure}

\begin{figure}
\centering
\includegraphics[width=\columnwidth]{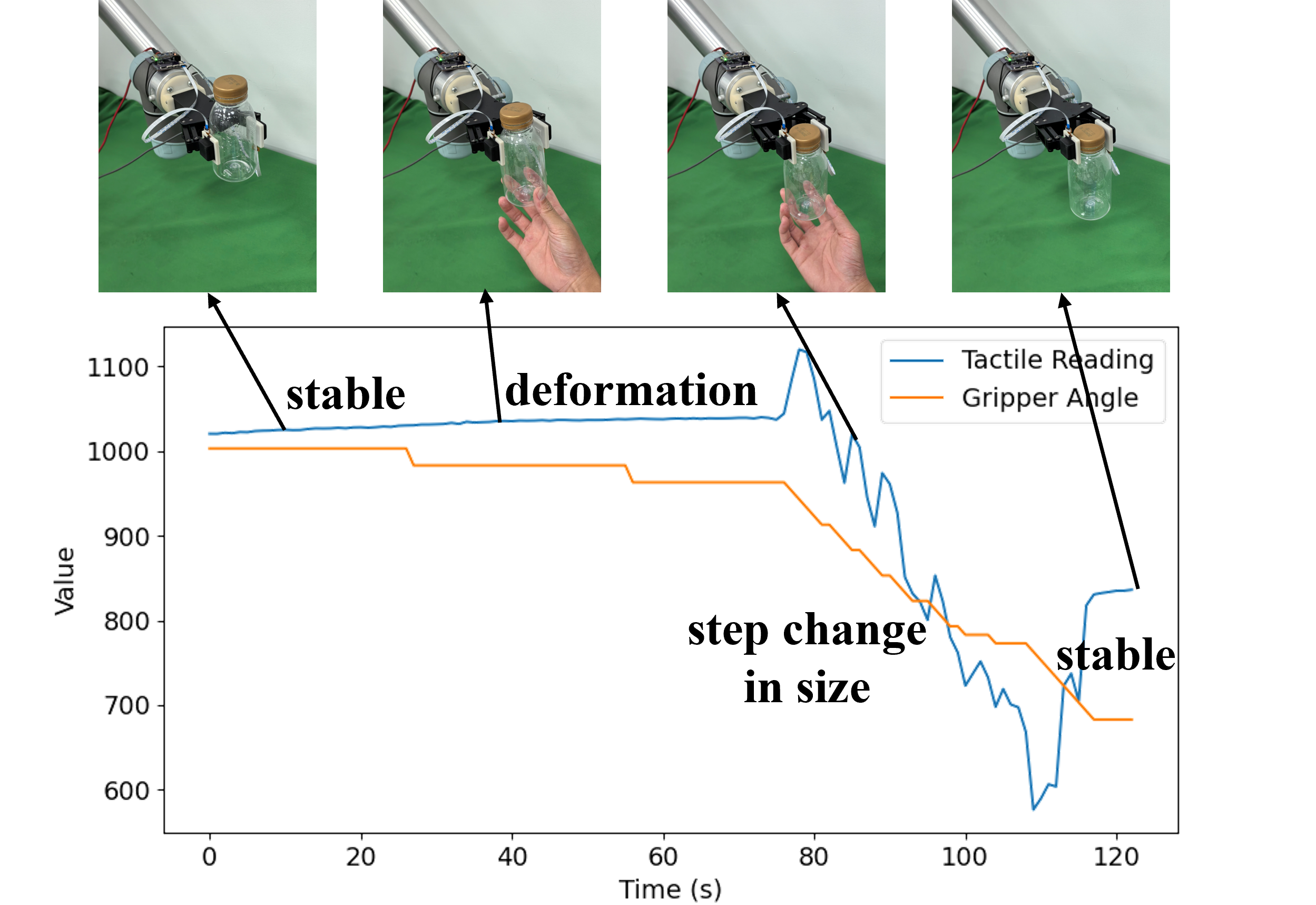}
\caption{A snapshot for the experiment on milk bottle. A human is trying to pull the bottle downwards from the bottom of the bottle and fingers starts to adapt its location.} 
\label{applyforce}
\end{figure}

We employ the deep neural network to learn the adaptive grasp skill. For the initial grasp generator, we use a neural network model with a learning rate of 0.001, a loss function based on cross-entropy, the optimizer SGD, a batch size of 64 and 50 epoches. The dataset is divided into training and validation sets at an 8:2 ratio. We use GMM to train the stability estimator, with the number of classes set to 2, the maximum number of iterations set to 100, and the initialization method chosen as K-means clustering. For training the grasp adaptation strategy, we chose Adam as the optimizer with a learning rate set to 0.001, a batch size of 64, and 100 epochs. The loss function used is Mean Squared Error (MSE).

\subsection{Grasp Operation on Novel Objects}
In order to simulate the uncertainty in the grasping process, we chose two approaches: increasing the mass of the object by adding water and artificially applying disturbing forces and vibrations.

\begin{figure}
\centering
\includegraphics[width=\columnwidth]{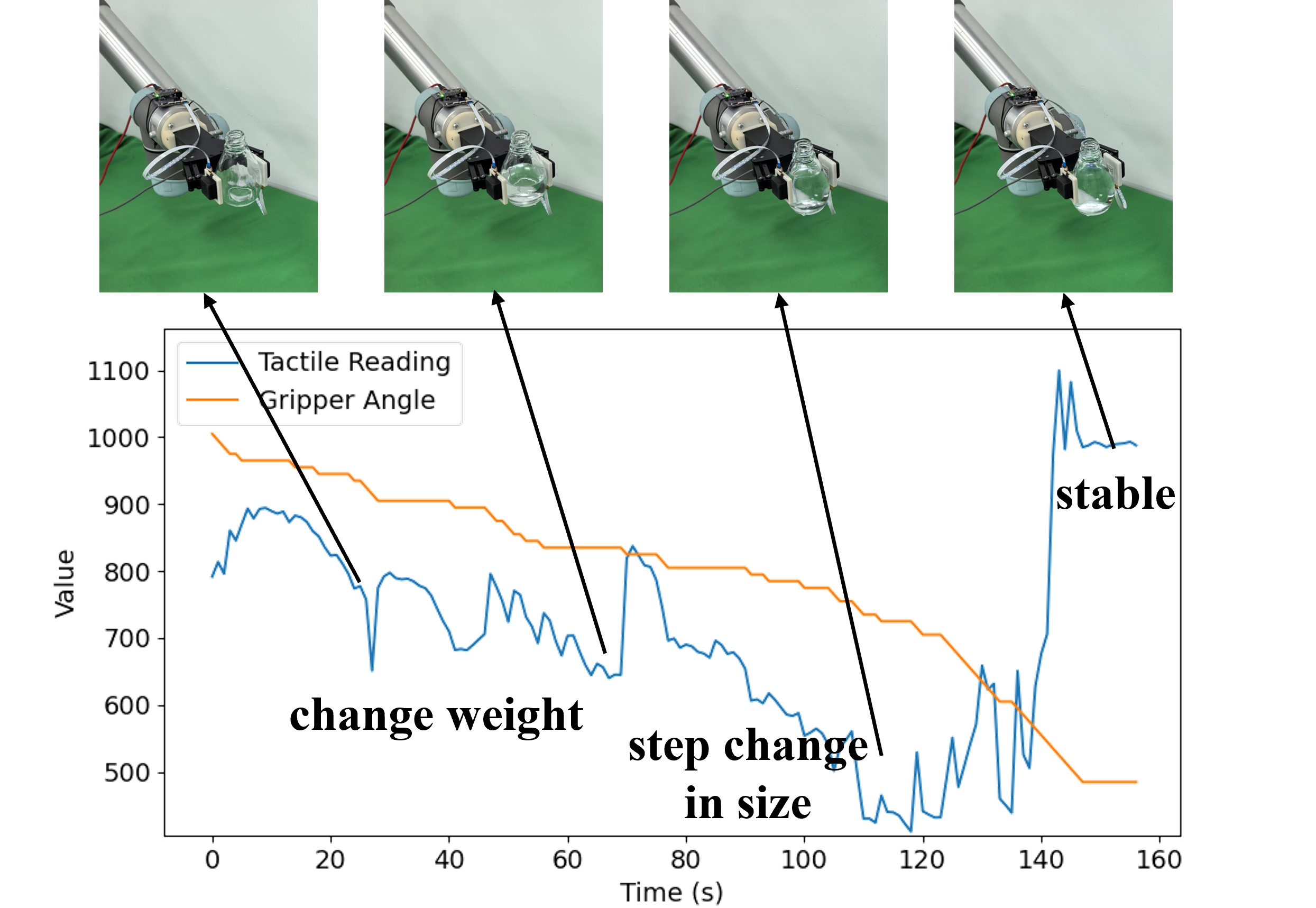}
\caption{A snapshot for the experiment on wine bottle. A snapshot for the experiment on wine bottle. A human is trying to pour water into the bottle and fingers starts to adapt its location.} 
\label{addwater}
\end{figure}

To validate the effectiveness of our grasp adaptation strategy, we chose a variety of novel objects including milk bottles, medicine jars, perfume, etc., as shown in Fig. \ref{testitems} for our tests. Our validation method is to first generate a stable initial grip given by generator. When the object weights are changing by adding water or disturbed by a human, the grasp adaptation strategy is triggered to keep the grasp stable by varying the gripper angle.

 To verify the effectiveness of the grasping adaptation method on different objects in a real robot hand, the weight of the object changes due to the addition of water or affected by human interference. In this experiment, initial grasp remains stable and artificial perturbations or forces are applied to the object to simulate external disturbances such as vibration, slippage, and changes in the object's mass. 
 
Taking the milk bottle as an example in Fig. \ref{applyforce}, when different levels of force are applied to the bottle, the gripper adapts its angle to maintain stability. In the case of grasping the wine bottle, see Fig. \ref{addwater}. We interfered by adding water to the bottle to continuously increase its mass. When an increase in mass destabilizes the bottle, the gripper adapts its configuration to maintain stability.

\begin{table}[]
\caption{THE COMPARISON OF THE SUPPORTED OBJECT WEIGHTS
(WITH VS. W/O GRASP ADAPTATION)}
\label{tab:my-table}
\begin{tabular}{|l|c|c|}
\hline
\textbf{Object} & \multicolumn{1}{l|}{\textbf{Max Weight Without (g)}} & \multicolumn{1}{l|}{\textbf{Max Weight With (g)}} \\ \hline
pillbox         & 34                                                   & 55                                                \\ \hline
tea can         & 45                                                   & 72                                                \\ \hline
mouthwash       & 126                                                  & 177                                               \\ \hline
milk bottle     & 144                                                  & 189                                               \\ \hline
wine bottle     & 149                                                  & 214                                               \\ \hline
perfume         & 94                                                   & 133                                               \\ \hline
ink             & 72                                                   & 129                                               \\ \hline
\end{tabular}
\end{table}

\section{Conclusion}
In this paper, a new adaptive grasping framework is proposed under the physical uncertainty of grasping objects. We first plan the initial grasping using behavior cloning. Secondly, Gaussian mixture model is used to learn the grasp stability bounds. During grasp execution, adaptive grasp strategy is triggered to maintain the stability. The effectiveness of the method is verified on a electric gripper with tactile sensors. 

In our method, the electric gripper shows limited adaptability when facing targets with abrupt shape changes, which sometimes leads to unstable grasps due to either the loss of contact during jump or the large impact force during contact. Future work will address this issue by focusing on optimizing the learning parameters in the grasping strategy and control algorithms. The experimental results are shown in Table \ref{tab:my-table}, demonstrating the maximum weight that can be stably grasped with and without the strategy.

 \section{Acknowledgment}
 This work was supported by the Fundamental Research Funds for the Central Universities under the grant agreement number 2042023kf0110,the Interdisciplinary Innovative Talents Foundation from Renmin Hospital of Wuhan University under the grant agreement number of JCRCYR-2022-002 and Wuhan key research and development project under the grant agreement number 2024060788020073.

% \addtolength{\textheight}{-12cm}   % This command serves to balance the column lengths
                                  % on the last page of the document manually. It shortens
                                  % the textheight of the last page by a suitable amount.
                                  % This command does not take effect until the next page
                                  % so it should come on the page before the last. Make
                                  % sure that you do not shorten the textheight too much.

%%%%%%%%%%%%%%%%%%%%%%%%%%%%%%%%%%%%%%%%%%%%%%%%%%%%%%%%%%%%%%%%%%%%%%%%%%%%%%%%

%%%%%%%%%%%%%%%%%%%%%%%%%%%%%%%%%%%%%%%%%%%%%%%%%%%%%%%%%%%%%%%%%%%%%%%%%%%%%%%%

%%%%%%%%%%%%%%%%%%%%%%%%%%%%%%%%%%%%%%%%%%%%%%%%%%%%%%%%%%%%%%%%%%%%%%%%%%%%%%%%

%%%%%%%%%%%%%%%%%%%%%%%%%%%%%%%%%%%%%%%%%%%%%%%%%%%%%%%%%%%%%%%%%%%%%%%%%%%%%%%%
\addtolength{\textheight}{-5cm}

\bibliographystyle{IEEEtran}
\bibliography{IEEEfull}

\end{document}